\documentclass[runningheads,orivec]{llncs}
\usepackage[T1]{fontenc}

\usepackage{caption}
\usepackage{subcaption}
\usepackage{subfig}

\usepackage{amssymb}
\usepackage{amsmath}
\usepackage{algorithm}
\usepackage{algorithmic}
\usepackage{booktabs}
\usepackage{multirow}
\usepackage{graphicx}
\usepackage{tikz}
\usepackage{longtable}
\usepackage{adjustbox}
\usepackage{makecell}
\usepackage{colortbl}
\usepackage{adjustbox}
\usepackage{float}
\usetikzlibrary{arrows.meta,calc,positioning,shapes.misc}
\usepackage[utf8]{inputenc}
\usepackage{textgreek}
\usepackage{url}
\usepackage{caption}
\usepackage{subcaption}
\usepackage{appendix}

\usepackage{booktabs}
\usepackage{threeparttable}
\usepackage[table]{xcolor}
\usepackage{array}
\usepackage{ragged2e}
\usepackage{adjustbox}
\usepackage{microtype}

\newcolumntype{L}[1]{>{\RaggedRight\arraybackslash}p{#1}}

\begin{document}



\title{Anarchy in the swarm: Testing informed and uninformed diversity-enhancing mechanisms within PSO framework}
\titlerunning{Anarchy in the swarm}

\author{{Piotr Urbańczyk\inst{1,2} \orcidID{0000-0001-8838-2354} \and
Aleksandra Urbańczyk\inst{1}\orcidID{0000-0002-6040-554X} }}
%
%
\institute{AGH University of Krakow, al.\ Mickiewicza 30, 30-059 Kraków, Poland
 \email{\{purbanczyk,aurbanczyk\}@agh.edu.pl}
\and
Jagiellonian University,
 ul.\ Gołębia 24, 31-007 Kraków, Poland
 \email{piotr.urbanczyk@uj.edu.pl}
}

\maketitle

\begin{abstract}
Particle Swarm Optimization (PSO) frequently suffers from premature convergence. This paper introduces a family of  problem-informed diversity-enhancing strategies  that manipulate the swarm's social and cognitive components. These include opposing-best strategies that repel particles from optimal regions, negative learning strategies that guide exploration toward poor solutions, and reverse learning strategies that push particles away from inferior regions.
These socio-cognitive mechanisms 
are evaluated against an analogous suite of problem-unaware, explicit randomization strategies that inject randomness either into velocity update components or directly into position updates.
The results reveal that the effectiveness of diversity enhancement is determined primarily by how it is embedded within the swarm dynamics, rather than by the mere presence of extraneous problem-informed guidance.
Particularly, random perturbations introduced at the velocity-update level consistently outperform those applied directly to particle positions.
\end{abstract}


\section{Introduction}
\label{sec:introduction}
Introduced by Kennedy and Eberhart in 1995 \cite{kennedy1995particle,eberhart1995new}, PSO draws inspiration from the social behavior and coordinated movement of animal formations such as bird flocks and fish schools. The algorithm operates by maintaining a population of candidate solutions, named particles, which navigate through the search space by adjusting their positions based on both ``collective knowledge'' (social component) and ``individual experience'' (cognitive component). Each particle is influenced by the global best position discovered by the entire swarm and retains memory of its personal best position. Through iterative updates governed by simple mathematical rules, the swarm collectively explores the problem landscape in order to converge toward near-optimal solutions.

Despite its widespread success and recognition, standard PSO suffers from a well-documented limitation: a tendency toward premature convergence—particularly when confronted with complex, multimodal, high-dimensional optimization problems. This phenomenon occurs when particles rapidly cluster around suboptimal solutions, effectively becoming trapped in local minima before adequately exploring the search space.
This vulnerability stems from the structural formulation of standard PSO, which heavily prioritizes exploitation. Because the velocity update strictly binds particles to their historical bests, the swarm rapidly loses spatial diversity. Once the population collapses around a promising region, it lacks the kinetic energy required to escape local basins of attraction, permanently trapping the search process in suboptimal solutions.

\textls[-10]{These drawbacks have driven extensive research into PSO variants, modifications, and hybrid algorithms, resulting in a vast literature of enhanced PSO-based methods
\cite{shi1998modified,kennedy1999small,higashi2003gaussian,thangaraj2011particle,esmin2013hybrid,zhang2015comprehensive,sengupta2019particle,jain2022Overview,minku2023introduction,abualigah2025particle,chauhan2025learning,urbanczyk2025sequential}.
This paper contributes to that body of work, providing modified PSO variants that explicitly address above-mentioned issues through novel learning and diversity-enhancing strategies.
The proposed approach is based on the premise that both high-quality and poor solutions provide valuable information about the fitness landscape and can serve as reference points for guided exploration. This perspective aligns with Albert Bandura’s social learning theory, which states that individuals acquire knowledge not only through their own experiences but also by observing the behavior of others, which allows them to imitate successful strategies and avoid the mistakes made by others \cite{bandura1986}. Similar principles have proven effective in population-based optimization, inspiring socio-cognitive operators in evolutionary strategies \cite{iccs2021,urbanczykes2024} and interaction mechanisms in agent-based island evolutionary algorithms \cite{tbo}.}

The paper defines a family of three problem-informed strategies: repulsion from best positions, attraction toward worst positions (negative learning strategies), and repulsion from worst positions (reverse learning strategies).
Each behavioral category is instantiated through two algorithmic variants that apply the core mechanism to different components of the PSO velocity update mechanism—specifically, the social component (based on global best/worst) and the cognitive component (based on personal best/worst).
This palette is complemented by a corresponding family of uninformed or problem-unaware explicit randomization strategies that introduce controlled stochasticity into particle dynamics, affecting either one or both components of the velocity update or the position update mechanism.

To rigorously evaluate these two families of strategies, this paper presents a comprehensive empirical study on a suite of 32 benchmark functions encompassing diverse landscape characteristics. All experiments are conducted at challenging dimensionalities of 100, 500, and 1000 variables.

\section{Related Work}
\label{sec:related}
\enlargethispage{1.5\baselineskip}
Several conceptually related ideas can be found in the PSO literature.
Early work such as Multi-Phase PSO (MPPSO) \cite{alkazemi2000multiphase,alkazemi2002multiphase} divided the swarm into sub-swarms alternating between attraction and repulsion phases, explicitly reversing social component of the velocity update. Two Sub-Swarms PSO (TSPSO) \cite{chen2005twosubswarms} similarly assigned opposing search directions to separate groups, periodically exchanging particles to maintain diversity. Attractive-and-Repulsive PSO (ARPSO) \cite{riget2002arpso} implemented swarm-wide phase switching triggered by diversity thresholds.

\textls[-20]{A more counterintuitive extension involves learning from inferior solutions. Reverse-Learning PSO (RLPSO) \cite{xia2014reverselearning} allows historical worst particles to serve as conditional attractors, particularly under stagnation. Niching-based Reverse PSO (NRPSO) \cite{dong2018reverselearning} refines this idea using clustering and simulated annealing to promote exploration of low-fitness niches.
A distinct research direction employs worst-position memory in a genuinely repulsive manner. Reverse Direction Supported PSO (RDS-PSO) \cite{comak2016generalized} introduces parameters that balance attraction toward best positions with repulsion from personal and global worst positions, spanning canonical PSO and fully reverse-directed dynamics as limiting cases. Similarly, PSO with Avoidance of Worst Locations (PSO-AWL) \cite{mason2016avoidance} and Parallel Global Best–Worst PSO (GBWPSO) \cite{kumar2023parallel} explicitly steer particles away from known poor regions, biasing trajectories toward higher-quality areas while preserving convergence pressure.}

Perturbation constitutes another major diversity-enhancement strategy. Inspired by mutation in evolutionary algorithms, early hybrids incorporated Gaussian mutation into standard PSO \cite{higashi2003gaussian,esmin2013hybrid}. Gaussian Particle Swarm Optimization (GPSO) \cite{secrest2003gaussian} introduced truncated Gaussian shifts around best positions, while adaptive Bare-Bones PSO variants \cite{zhang2014adaptive} dynamically adjusted particle-specific variance according to convergence degree and swarm diversity. Heavy-tailed and adaptive perturbations further strengthen global search capability. Cauchy-based mutations \cite{stacey2003mutation,wang2007cauchy,krohling2009barebones} enable long jumps that facilitate escape from complex local basins, though sometimes at the cost of unimodal efficiency. Lévy-flight PSO \cite{hakli2014levy} and chaos-based perturbation mechanisms \cite{yong2015adaptive} introduce controlled large redistributions under stagnation. Recent adaptive hybrids such as AGMPSO \cite{chen2021adaptive} and MDE-DPSO \cite{xu2025hybrid} combine Gaussian disturbance, mutation, and differential evolution operators under dynamic control strategies.

\section{Proposed Diversity-Enhancing Strategies}
\label{sec:methods}
\enlargethispage{1.5\baselineskip}
Formally, PSO operates on a population of candidate solutions, called a swarm \(P\), that explore a \(d\)-dimensional search space. Each particle $X_i$ is defined by its position \(\vec{x_i} = (x_{i}^{1}, x_{i}^{2}, \dots, x_{i}^{d})\), velocity \(\vec{v_i} =\allowbreak\ (v_{i}^{1}, v_{i}^{2},\dots, v_{i}^{d})\), and its personal best-found position \(\vec{b_i} = (b_{i}^{1}, b_{i}^{2}, \dots, b_{i}^{d}) \).
The swarm's evolution is guided by updating each particle's velocity and position at each iteration. The velocity update directs the particle towards its personal best (\(\vec{b_i}\)) and the global best (\(\vec{b_g}\)) position found by the entire swarm:
\begin{gather}
\vec{v_{i}}' = \underbrace{\omega \vec{v}_{i}}_{\text{inertia}} + 
\underbrace{c_1 r_1 (\vec{b}_{i} - \vec{x}_{i})}_{\text{cognitive component}} + 
\underbrace{c_2 r_2 (\vec{b}_{g} - \vec{x}_{i})}_{\text{\ \ \text{social component}\ \ }}
\label{eq:velocity_update}
\\
\vec{x_{i}}' = \vec{x}_{i} + \vec{v_{i}}'
\label{eq:position_update}
\end{gather}
Here, \(\omega\) is the inertia weight, \(c_1\) and \(c_2\) are acceleration coefficients (often called \textit{cognitive} and \textit{social} coefficients, respectively), and \(r_1, r_2\) are random values in \([0, 1]\). The position is then updated by the resulting velocity.
After each move, the particle's new position is evaluated using the objective function \(f\). The personal and global best positions are updated if a better solution is found.
After each particle update, feasibility is enforced by component-wise clipping, i.e., each coordinate that exceeds the lower or upper bound of the search space is projected back to the nearest feasible value.
This iterative process continues until a termination criterion is met, typically based on maximum iterations, function evaluations, or convergence thresholds.

\subsection{Informed Strategies}

Problem-informed swarm diversity enhancement strategies are organized into three fundamental behavioral categories, each defining two  particle roles and instantiated through corresponding algorithmic variants. All proposed algorithms extend the PSO framework by introducing a subswarm of particles assigned to one of these roles. The rest of the particles behave normally, guided by Eq.~\eqref{eq:velocity_update} and Eq.~\eqref{eq:position_update}, as in ordinary PSO. A comprehensive comparison of the introduced roles and strategies is provided in Table~\ref{tab:strategies_comparison}.

\subsubsection{Repulsion Strategies}

RebelPSO and RejectorPSO belong to the family of strategies that introduce exploratory pressure by deliberately repelling particles from known optimal regions, counteracting PSO's natural convergence tendency. By inverting the standard attraction forces, these mechanisms force particles to explore alternative areas of the search space.
\begin{gather}
\vec{v_{i}}' = \omega \vec{v}_{i} + c_1 r_1 (\vec{b}_{i} - \vec{x}_{i}) + c_3 r_3 (\vec{x}_{i} - \vec{b}_{g})
\label{eq:velocity_update_rebel}
\\
\vec{v_{i}}' = \omega \vec{v}_{i} + c_4 r_4 (\vec{x}_{i} - \vec{b}_{i}) + c_2 r_2 (\vec{b}_{g} - \vec{x}_{i})
\label{eq:velocity_update_rejector}
\end{gather}
\textls[-7]{RebelPSO assigns a fixed fraction of the swarm, $\rho_3$, to the role of ``rebels''. These particles preserve the standard inertia and cognitive components but invert the social term, resulting in repulsion from the global best position~$\vec{b_g}$. Their velocity is updated according to Eq.~\eqref{eq:velocity_update_rebel}, where $c_3$ denotes the corresponding acceleration coefficient (analogously, coefficients $c_4$–$c_8$ parameterize the modified components in the remaining strategy variants).
RejectorPSO follows a similar principle, but applies repulsion to the cognitive component. A fraction $\rho_4$ of particles is designated as ``rejectors'', which are driven away from their personal best position~$\vec{b_i}$. This mechanism encourages abandonment of previously exploited regions in favor of renewed exploration. The velocity update of rejector particles is defined by Eq.~\eqref{eq:velocity_update_rejector}.
}

\enlargethispage{2.5\baselineskip}
\subsubsection{Negative Learning Strategies}

This family of strategies implements a form of negative learning by intentionally guiding particles toward the worst solutions encountered. This counterintuitive mechanism leverages the information content of failures to identify potentially promising unexplored regions. To formalize variants incorporating worst-position memory, we extend the standard PSO formulation. The particle representation is augmented
with
$\vec{w}_i$ that denotes the particle’s personal worst position. In addition to the global best position $\vec{b}_g$, the swarm also keeps track of a global worst position $\vec{w}_g$.
\begin{gather}
\vec{v_{i}}' = \omega \vec{v}_{i} + c_1 r_1 (\vec{b}_{i} - \vec{x}_{i}) + c_5 r_5 (\vec{w}_{g} - \vec{x}_{i})
\label{eq:velocity_update_contrarian}
\\
\vec{v_{i}}' = \omega \vec{v}_{i} + c_6 r_6 (\vec{w}_{i} - \vec{x}_{i}) + c_2 r_2 (\vec{b}_{g} - \vec{x}_{i})
\label{eq:velocity_update_defeatist}
\end{gather}
ContrarianPSO designates a fraction $\rho_5$ of the swarm as ``contrarians.'' These particles replace the standard social component with an attraction toward the global worst position $\vec{w_g}$. The velocity update mechanism for a contrarian particle is expressed by Eq.~\eqref{eq:velocity_update_contrarian}.
DefeatistPSO applies the negative learning principle to the cognitive component. A fraction $\rho_6$ of the swarm is designated as ``defeatists,'' which are attracted to their own personal worst position $\vec{w_i}$. The velocity update for a defeatist particle is presented in Eq.~\eqref{eq:velocity_update_defeatist}.

\subsubsection{Reverse Learning Strategies}

This family of strategies uses inferior solutions as reference points to guide exploration away from demonstrably poor regions, creating a complementary exploratory dynamic.
\begin{gather}
\vec{v_{i}}' = \omega \vec{v}_{i} + c_1 r_1 (\vec{b}_{i} - \vec{x}_{i}) + c_7 r_7 (\vec{x}_{i} - \vec{w}_{g})
\label{eq:velocity_update_eschewer}
\\
\vec{v_{i}}' = \omega \vec{v}_{i} + c_8 r_8 (\vec{x}_{i} - \vec{w}_{i}) + c_2 r_2 (\vec{b}_{g} - \vec{x}_{i})
\label{eq:velocity_update_escapist}
\end{gather}
EschewerPSO designates a fraction $\rho_7$ of the swarm as ``eschewers.'' These particles are repelled from the global worst position $\vec{w_g}$. The velocity update for an eschewer particle is Eq.~\eqref{eq:velocity_update_eschewer}.
EscapistPSO applies the same principle to the cognitive component. A fraction $\rho_8$ of particles are ``escapists,'' repelled from their personal worst position $\vec{w_i}$. The velocity update dynamic is defined by Eq.~\eqref{eq:velocity_update_escapist}.

\begin{table}[htbp]
\centering
\footnotesize
\setlength{\tabcolsep}{6pt}        
\renewcommand{\arraystretch}{1.15} 
\begin{threeparttable}
\caption{Comparison of novel particle roles introduced in proposed strategies.}
\label{tab:strategies_comparison}

\begin{tabular}{@{}L{0.18\linewidth} L{0.40\linewidth} L{0.15\linewidth} L{0.21\linewidth}@{}}
\toprule
\textbf{Particle role} & \textbf{Modification} & \textbf{Direction} & \textbf{Target} \\
\midrule

\rowcolor{gray!50}
\multicolumn{4}{@{}l@{}}{\textbf{Informed strategies}}\\

\rowcolor{gray!12}
\multicolumn{4}{@{}l@{}}{\textbf{Repulsion strategies}}\\
rebel    & social component    & away from & global best \\
rejector & cognitive component & away from & individual best \\
\addlinespace[3pt]

\rowcolor{gray!12}
\multicolumn{4}{@{}l@{}}{\textbf{Negative learning strategies}}\\
contrarian & social component    & toward & global worst \\
defeatist  & cognitive component & toward & individual worst \\
\addlinespace[3pt]

\rowcolor{gray!12}
\multicolumn{4}{@{}l@{}}{\textbf{Reverse learning strategies}}\\
eschewer & social component    & away from & global worst \\
escapist & cognitive component & away from & individual worst \\

\rowcolor{gray!50}
\multicolumn{4}{@{}l@{}}{\textbf{Uninformed strategies}}\\

anarchic & social component    & random & n.a. \\
amnesiac & cognitive component & random & n.a. \\
erratic  & both velocity update components & random & n.a. \\ 
wanderer & added velocity update component & random & n.a. \\ 
drifter  & added position update component & random & n.a. \\

\bottomrule
\end{tabular}
\end{threeparttable}
\end{table}

\subsection{Uninformed strategies}
\enlargethispage{2.5\baselineskip}
\textls[-10]{Analogously, AnarchicPSO and AmnesiacPSO are extensions of the standard PSO algorithm, enhanced with a behavioral twist through assigning roles to subswarms of particles: ``anarchic'' particles ignore the global best and instead move in a random direction. The velocity update mechanism of such particles is governed by Eq.~\eqref{eq:velocity_update_anarchic}. The random direction vector is scaled by a parameter $\lambda$. ``Amnesiac'' preserves inertia and social components but replaces the cognitive component (attraction to personal best) with a  (scaled) random vector as in Eq.~\eqref{eq:velocity_update_amnesiac}.}
    
In a similar manner, in ErraticPSO an ``erratic'' fraction of solution candidates updates their velocity based only on inertia and a random vector, ignoring personal and global best attractors. Within WandererPSO a subswarm of ``wanderer'' particles adds a scaled random vector on top of their standard velocity. The movement of these particles is defined by Eq.~\eqref{eq:velocity_update_erratic} and~\eqref{eq:velocity_update_wanderer}, respectively.
\begin{gather}
\vec{v_{i}}' = \omega \vec{v}_{i}
c_1 r_1 (\vec{b}_{i} - \vec{x}_{i})
+ \lambda \vec{\xi}_i,
\qquad \vec{\xi}_i \sim U(-1,1)^d
\label{eq:velocity_update_anarchic}
\\
\vec{v_{i}}' = \omega \vec{v}_{i}
+ \lambda \vec{\xi}_i +
c_2 r_2 (\vec{b}_{g} - \vec{x}_{i})
\qquad \vec{\xi}_i \sim U(-1,1)^d
\label{eq:velocity_update_amnesiac}
\\
\vec{v_{i}}' = \omega \vec{v}_{i}
+ \lambda \xi_{ij},
\qquad \vec{\xi}_i \sim U(-1,1)^d
\label{eq:velocity_update_erratic} 
\\
\vec{v_{i}}' = \omega \vec{v}_{i} +
c_1 r_1 (\vec{b}_{i} - \vec{x}_{i}) +
c_2 r_2 (\vec{b}_{g} - \vec{x}_{i})
+ \lambda \vec{\xi}_i,
\qquad \vec{\xi}_i \sim U(-1,1)^d
\label{eq:velocity_update_wanderer} 
\end{gather}

DrifterPSO also designates the subswarm of ``drifter'' particles. Their velocity update mechanism remains identical to the standard PSO formulation. The core change is that the Gaussian noise term $\epsilon_{ij}$ is integrated directly into the position update as in Eq.~\eqref{eq:position_update_drifter}.
\begin{equation}
    \vec{x_{i}}' = \vec{x}_{i} + \vec{v_{i}}' + \vec{\epsilon}_i
\qquad \vec{\epsilon}_i \sim \mathcal{N}(\vec{0}, \sigma^2 I)
\label{eq:position_update_drifter} 
\end{equation}

\begin{figure}[H]
    \centering
    \begin{subfigure}{0.48\textwidth}
        \centering
        \resizebox{\linewidth}{!}{%
            \begin{tikzpicture}[
                point_style/.style={circle, fill=black, inner sep=1.5pt},
                vector_style/.style={-Latex, thick},
                label_style/.style={font=\small},
                component_label_style/.style={font=\footnotesize, sloped, midway},
                guide_line_style/.style={dashed, gray!60, thin}
            ]
            \coordinate (xi_t) at (0,0);
            \coordinate (pbest_i) at (2.2,2.7);
            \coordinate (gbest) at (8,0.5);
            \coordinate (inertia_vec) at (0.9, -1.2);
            \coordinate (cognitive_vec_raw) at ($(pbest_i) - (xi_t)$);
            \coordinate (cognitive_vec) at ($1.4*(cognitive_vec_raw)$);
            \coordinate (social_vec_raw) at ($(gbest) - (xi_t)$);
            \coordinate (social_vec) at ($0.5*(social_vec_raw)$);
            \coordinate (xi_vec) at (0.7, -1.3);                 
            \coordinate (lambda_xi_vec) at ($0.8*(xi_vec)$);    
            \coordinate (v_tip1) at ($(xi_t) + (inertia_vec) + (cognitive_vec) + (social_vec) + (lambda_xi_vec)$);
            \coordinate (xi_t_plus_1) at (v_tip1);
            \node[point_style, label={[label_style]below left:$x_i$}] at (xi_t) {};
            \node[draw, cross out, inner sep=1.5pt, label={[label_style]above:$b_{i}$}] at (pbest_i) {};
            \node[draw, cross out, inner sep=1.5pt, label={[label_style]below right:$b_{g}$}] at (gbest) {};
            \node[circle, fill=black, inner sep=1.5pt, label={[label_style]right:$x_i'$}] at (xi_t_plus_1) {};
            \draw[guide_line_style] (xi_t) -- (pbest_i);
            \draw[dash dot, gray!60, thin] (xi_t) -- (gbest);
            \draw[vector_style, ultra thick, opacity=0.5] (xi_t) -- node[component_label_style, below, near end] {${v}_i'$} (v_tip1);
            \draw[vector_style, densely dotted] (xi_t) -- node[component_label_style, above] {$\omega v_{i}$} ($(xi_t) + (inertia_vec)$) coordinate (tip_inertia);
            \draw[vector_style, dashed] (tip_inertia) -- node[component_label_style, above, near end] {$c_1 r_1 (b_{i} - x_{i})$} ($(tip_inertia) + (cognitive_vec)$) coordinate (tip_cognitive);
            \draw[vector_style, dash dot] (tip_cognitive) -- node[component_label_style, above] {$c_2 r_2 (b_{g} - x_{i})$} ($(tip_cognitive) + (social_vec)$) coordinate (tip_social);
            \draw[vector_style, loosely dashed] (tip_social) --  node[component_label_style, above] {$\lambda \xi_i$}
              ($(tip_social) + (lambda_xi_vec)$) coordinate (tip_noise);
            \end{tikzpicture}%
        }
        \caption{Wanderer particle's position update.}
        \label{fig:PSO_geometric_illustration}
    \end{subfigure}
    \hfill 
    \begin{subfigure}{0.48\textwidth}
        \centering
        \resizebox{\linewidth}{!}{%
            \begin{tikzpicture}[
                point_style/.style={circle, fill=black, inner sep=1.5pt},
                vector_style/.style={-Latex, thick},
                label_style/.style={font=\small},
                component_label_style/.style={font=\footnotesize, sloped, midway},
                guide_line_style/.style={dashed, gray!60, thin}
            ]
            \coordinate (xi_t) at (0,0);
            \coordinate (pbest_i) at (2.2,2.7);
            \coordinate (gbest) at (8,0.5);
            \coordinate (inertia_vec) at (0.9, -1.2);
            \coordinate (cognitive_vec_raw) at ($(pbest_i) - (xi_t)$);
            \coordinate (cognitive_vec) at ($1.4*(cognitive_vec_raw)$);
            \coordinate (social_vec_raw) at ($(gbest) - (xi_t)$);
            \coordinate (social_vec) at ($0.5*(social_vec_raw)$);
            \coordinate (v_tip1) at ($(xi_t) + (inertia_vec) + (cognitive_vec) + (social_vec)$);
            \coordinate (xi_t_plus_1) at (v_tip1);
            \node[point_style, label={[label_style]below left:$x_i$}] at (xi_t) {};
            \node[draw, cross out, inner sep=1.5pt, label={[label_style]above:$b_{i}$}] at (pbest_i) {};
            \node[draw, cross out, inner sep=1.5pt, label={[label_style]below right:$b_{g}$}] at (gbest) {};
            \draw[guide_line_style] (xi_t) -- (pbest_i);
            \draw[dash dot, gray!60, thin] (xi_t) -- (gbest);
            \draw[vector_style, ultra thick, opacity=0.5] (xi_t) -- node[component_label_style, below, near end] {${v}_i'$} (v_tip1);
            \draw[vector_style, densely dotted] (xi_t) -- node[component_label_style, above] {$\omega v_{i}$} ($(xi_t) + (inertia_vec)$) coordinate (tip_inertia);
            \draw[vector_style, dashed] (tip_inertia) -- node[component_label_style, above, near end] {$c_1 r_1 (b_{i} - x_{i})$} ($(tip_inertia) + (cognitive_vec)$) coordinate (tip_cognitive);
            \draw[vector_style, dash dot] (tip_cognitive) -- node[component_label_style, above] {$c_2 r_2 (b_{g} - x_{i})$} ($(tip_cognitive) + (social_vec)$) coordinate (tip_social);
            \def\sigmaPerturb{0.7}; 
            \coordinate (xi_t_plus_1_perturbed_final) at ($(xi_t_plus_1) + (rand*0.3, rand*0.3)$);
            \draw[black, dashed, thin] (xi_t_plus_1) circle (\sigmaPerturb);
            \foreach \i in {1,...,95} { 
                \pgfmathsetmacro{\randnormx}{(rand+rand+rand+rand+rand+rand)/2 * \sigmaPerturb * 0.4}
                \pgfmathsetmacro{\randnormy}{(rand+rand+rand+rand+rand+rand)/2 * \sigmaPerturb * 0.5}
                \node[circle, fill=gray!70, inner sep=0.1pt, opacity=0.5] at ($(xi_t_plus_1) + (\randnormx, \randnormy)$) {};
            }
            \coordinate (x_final) at ($(xi_t_plus_1) + (0.45, \sigmaPerturb-1.0)$);
            \coordinate (sigma_perimeter_right)  at ($(xi_t_plus_1) + (10:\sigmaPerturb)$);
            \node[circle, fill=black, inner sep=1.5pt, label={[label_style, yshift=1mm]below right:$x_i'$}] at (x_final) {};
            \draw[component_label_style, black] (xi_t_plus_1) -- node[component_label_style, black, above, midway] {$\sigma$} (sigma_perimeter_right);
            \end{tikzpicture}%
        }
        \caption{Drifter particle's position update.}
        \label{fig:PerturbationPSO_geometric_illustration}
    \end{subfigure}
    \caption{\textls[-37]{Two-dimensional geometric comparison of wanderer and drifter particle dynamics.}}
    \label{fig:pso_comparison}
\end{figure}
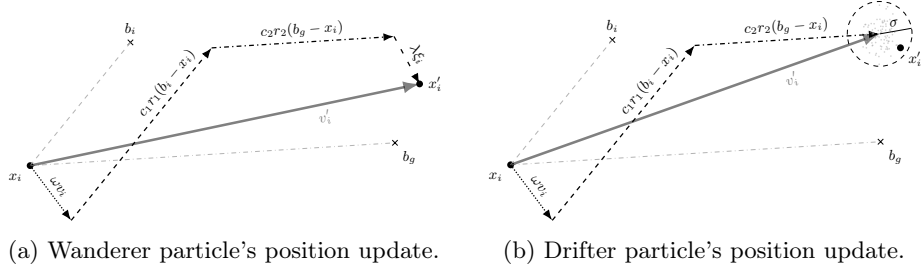

\section{Experiments and analysis}
\label{sec:experiments}
The algorithms were tested on a comprehensive suite of 32 scalable benchmark functions, selected to encompass a wide variety of problem features that challenge different aspects of optimization algorithms. The suite combines functions from the CEC (Congress on Evolutionary Computation) special sessions \cite{liang2013cec2013,wu2017cec2017} with classic test functions \cite{jamil2013survey}. This combination is motivated by studies showing that algorithms tailored to specific CEC suites may exhibit degraded performance on other problem types \cite{piotrowski2023benchmark}. The benchmark includes unimodal, multimodal, separable, non-separable, rotated, shifted, and composite functions with varying degrees of ill-conditioning.
All experiments were conducted in three dimensionalities: $d \in \{100, 500, 1000\}$.

\subsection{Experimental Setup}

\textls[-10]{In this study, PSO and all its variants were tuned systematically with the same procedure to ensure fair comparison. Parameter tuning employed the \texttt{irace} (iterated racing for automatic algorithm configuration) methodology and its associated R package \cite{lopezibanez2016irace}. 
Each algorithm was tuned with 1000 candidate configurations per parameter on three representative problems—Sphere, Rastrigin, and Ackley—covering diverse landscape characteristics \cite{los2018misfit}. Each 100-variable instance allowed 25,000 evaluations with 100 particles, tested over five independent runs. These tuning problems were excluded from the final benchmark suite.
Algorithms were implemented in the \texttt{jMetalPy} framework \cite{benitezhidalgo2019jmetalpy}.
The final experimental evaluation was conducted with a budget of 25,000 function evaluations and a swarm size of 100 particles, averaged over 50 independent runs per problem and dimensionality.
Parameter tuning and experiments were executed on the HPC cluster.
}

\subsection{Overview}
Table \ref{tab:best-algorithm-summary} reports the best-performing algorithm for each of the 32 benchmark problems across three dimensionalities. No single method dominates universally---with the exception of ErraticPSO, every variant appears at least once as the top performer. However, clear patterns emerge: WandererPSO, DefeatistPSO, DrifterPSO, and AmnesiacPSO occur most frequently among the best methods. Standard PSO appears only sporadically, indicating that diversity-enhancing modifications substantially improve baseline performance. Furthermore, the optimal algorithm frequently shifts with increasing dimensionality. In particular, WandererPSO and DefeatistPSO increasingly replace other methods as dimensionality grows, suggesting favorable scalability characteristics.

\begin{table}[ht!]
\centering
\caption{Best performing algorithm for each problem across all dimensionalities.}
\label{tab:best-algorithm-summary}
\resizebox{\textwidth}{!}{
\begin{tabular}{lccc}
\toprule
\textbf{Problem} & \textbf{d=100} & \textbf{d=500} & \textbf{d=1000} \\
\midrule
Alpine N1 & PSO & WandererPSO & WandererPSO \\
Crowned Cross & DrifterPSO & DrifterPSO & DrifterPSO \\
Egg-Holder & ContrarianPSO & PSO & PSO \\
Expanded Shaffer & DefeatistPSO & ContrarianPSO & ContrarianPSO \\
Generalized Schaffer N1 & DefeatistPSO & ContrarianPSO & WandererPSO \\
Generalized Schaffer N2 & AmnesiacPSO & WandererPSO & EscapistPSO \\
Generalized Schaffer N3 & DefeatistPSO & DefeatistPSO & DefeatistPSO \\
Generalized Schaffer N4 & EscapistPSO & EscapistPSO & PSO \\
Generalized Schmidt--Vetters & RebelPSO & RebelPSO & RebelPSO \\
Lennard-Jones Minimum Energy Cluster  & $\quad$ContrarianPSO$\quad$ & $\quad$ContrarianPSO$\quad$ & ContrarianPSO \\
Michalewicz & DrifterPSO & DrifterPSO & DrifterPSO \\
Mishra N3 & WandererPSO & AnarchicPSO & AnarchicPSO \\
Mishra N4 & WandererPSO & AnarchicPSO & AnarchicPSO \\
Modified Rosenbrock No.02 & AmnesiacPSO & DefeatistPSO & DefeatistPSO \\
Rotated Bent Cigar & DefeatistPSO & DefeatistPSO & DefeatistPSO \\
Rotated Discus & EschewerPSO & AmnesiacPSO & WandererPSO \\
Rotated High Conditioned Elliptic & PSO & WandererPSO & DefeatistPSO \\
Salomon & RebelPSO & WandererPSO & WandererPSO \\
Schwefel N20 & DefeatistPSO & DefeatistPSO & DefeatistPSO \\
Schwefel N36 & DrifterPSO & WandererPSO & WandererPSO \\
Schwefel N6 & DefeatistPSO & DefeatistPSO & AmnesiacPSO \\
Shifted Schwefel & AmnesiacPSO & AmnesiacPSO & AmnesiacPSO \\
Shifted and Rotated HGBat & WandererPSO & WandererPSO & DrifterPSO \\
Shifted and Rotated HappyCat & WandererPSO & WandererPSO & DrifterPSO \\
Shifted and Rotated Schaffer F7 & AmnesiacPSO & AmnesiacPSO & AmnesiacPSO \\
Shifted and Rotated Weierstrass & DrifterPSO & DrifterPSO & DrifterPSO \\
Shubert N3 & DrifterPSO & DrifterPSO & DrifterPSO \\
Shubert N4 & DrifterPSO & WandererPSO & WandererPSO \\
SineEnvelope & DrifterPSO & DefeatistPSO & DefeatistPSO \\
Stochastic & DefeatistPSO & DefeatistPSO & DefeatistPSO \\
StretchedV & AnarchicPSO & WandererPSO & PSO \\
Styblinski--Tang & EscapistPSO & DefeatistPSO & EscapistPSO \\
\bottomrule
\end{tabular}
}
\end{table}

Tables \ref{tab:algorithm-best-count} and \ref{tab:algorithm-worst-count} quantify performance consistency by reporting the frequency of best and worst outcomes. DefeatistPSO achieves the highest number of best performances overall (22), followed by WandererPSO (19) and DrifterPSO (17). However, the distribution of worst-case outcomes reveals substantial variability. DrifterPSO accumulates 55 worst rankings—by far the largest number---indicating highly unstable behavior despite frequent successes. In contrast, WandererPSO and AmnesiacPSO combine strong best-performance counts with relatively moderate worst-case frequencies. Importantly, the relative performance profiles remain consistent across dimensions.

Table \ref{tab:perf_and_stats_combined} provides a holistic evaluation through min–max normalized fitness averaged across all problems. WandererPSO achieves the best overall performance, with the lowest mean normalized score (0.227). AmnesiacPSO (0.271) and DefeatistPSO (0.305) also significantly outperform standard PSO (0.414). In contrast, DrifterPSO yields the poorest aggregate performance (0.686).
Demšar's many-to-one procedure revealed that WandererPSO achieved the strongest statistically significant improvement over PSO ($p = 2.02 \times 10^{-8}$), followed by AmnesiacPSO ($p = 8.88 \times 10^{-3}$) and DefeatistPSO ($p = 9.92 \times 10^{-3}$). All remaining variants failed to achieve statistical significance after Holm correction. While several algorithms exhibited non-significant performance improvements, others showed degradation relative to PSO.

\begin{table}[ht!]
\centering
\begin{minipage}[t]{0.49\textwidth}
\centering
\captionof{table}{Number of times each algorithm achieved the \textbf{best} performance across benchmark problems.}
\label{tab:algorithm-best-count}
\scriptsize
\setlength{\tabcolsep}{3pt}
\renewcommand{\arraystretch}{0.95}
\resizebox{\linewidth}{!}{%
\begin{tabular}{lcccc}
\toprule
\textbf{Algorithm} & \textbf{d=100} & \textbf{d=500} & \textbf{d=1000} & \textbf{Total} \\
\midrule
DefeatistPSO & \textbf{7} & 8 & \textbf{7} & \textbf{22} \\
WandererPSO & 4 & \textbf{9} & 6 & 19 \\
DrifterPSO & \textbf{7} & 4 & 6 & 17 \\
AmnesiacPSO & 4 & 3 & 3 & 10 \\
ContrarianPSO & 2 & 3 & 2 & 7 \\
PSO & 2 & 1 & 3 & 6 \\
AnarchicPSO & 1 & 2 & 2 & 5 \\
EscapistPSO & 2 & 1 & 2 & 5 \\
RebelPSO & 2 & 1 & 1 & 4 \\
EschewerPSO & 1 & 0 & 0 & 1 \\
\bottomrule
\end{tabular}%
}
\end{minipage}\hfill
\begin{minipage}[t]{0.49\textwidth}
\centering
\captionof{table}{Number of times each algorithm achieved the \textbf{worst} performance across benchmark problems.}
\label{tab:algorithm-worst-count}
\scriptsize
\setlength{\tabcolsep}{3pt}
\renewcommand{\arraystretch}{0.95}
\resizebox{\linewidth}{!}{%
\begin{tabular}{lcccc}
\toprule
\textbf{Algorithm} & \textbf{d=100} & \textbf{d=500} & \textbf{d=1000} & \textbf{Total} \\
\midrule
DrifterPSO & \textbf{17} & \textbf{19} & \textbf{19} & \textbf{55} \\
ContrarianPSO & 8 & 5 & 5 & 18 \\
AnarchicPSO & 2 & 2 & 1 & 5 \\
ErraticPSO & 2 & 0 & 3 & 5 \\
AmnesiacPSO & 1 & 0 & 2 & 3 \\
EscapistPSO & 1 & 1 & 1 & 3 \\
PSO & 1 & 1 & 1 & 3 \\
RebelPSO & 0 & 2 & 0 & 2 \\
RejectorPSO & 0 & 2 & 0 & 2 \\
\bottomrule
\end{tabular}%
}
\end{minipage}
\end{table}

\begin{table}[ht!]
\centering
\caption{\textls[-10]{Summary of average performance and many-to-one statistical comparison against the control PSO. Min-max normalized fitness is averaged across dimensions (lower is better). Post-hoc $p$-values are Holm--Bonferroni adjusted ($\alpha=0.05$).}}
\label{tab:perf_and_stats_combined}
\scriptsize
\setlength{\tabcolsep}{4pt}
\renewcommand{\arraystretch}{0.95}
\resizebox{\linewidth}{!}{%
\begin{tabular}{lcccc@{\hspace{10pt}}|@{\hspace{10pt}}c@{\hspace{10pt}}|@{\hspace{10pt}}cc}
\toprule
\textbf{Algorithm} & \textbf{d=100} & \textbf{d=500} & \textbf{d=1000} & \textbf{mean} & \textbf{$p$-value} & \textbf{vs PSO} \\
\midrule
WandererPSO  & \textbf{0.249} & \textbf{0.213} & \textbf{0.220} & \textbf{0.227} & $2.02 \times 10^{-8}$ $^{\ast}$ & Better \\
AmnesiacPSO  & 0.272 & 0.253 & 0.287 & 0.271 & $8.88 \times 10^{-3}$ $^{\ast}$ & Better \\
DefeatistPSO & 0.320 & 0.300 & 0.295 & 0.305 & $9.92 \times 10^{-3}$ $^{\ast}$ & Better \\
RebelPSO     & 0.374 & 0.360 & 0.377 & 0.370 & $1.00$                          & Better (ns) \\
EschewerPSO  & 0.386 & 0.398 & 0.413 & 0.399 & $4.82 \times 10^{-1}$           & Better (ns) \\
ContrarianPSO& 0.433 & 0.399 & 0.403 & 0.412 & $5.66 \times 10^{-1}$           & Worse (ns) \\
PSO          & 0.429 & 0.412 & 0.401 & 0.414 & ---                             & Control \\
RejectorPSO  & 0.414 & 0.419 & 0.409 & 0.414 & $1.00$                          & Worse (ns) \\
AnarchicPSO  & 0.427 & 0.413 & 0.440 & 0.426 & $1.21 \times 10^{-1}$           & Better (ns) \\
EscapistPSO  & 0.432 & 0.434 & 0.423 & 0.430 & $9.70 \times 10^{-1}$           & Worse (ns) \\
ErraticPSO   & 0.482 & 0.458 & 0.482 & 0.474 & $6.74 \times 10^{-2}$           & Worse (ns) \\
DrifterPSO   & 0.671 & 0.699 & 0.688 & 0.686 & $5.66 \times 10^{-1}$           & Worse (ns) \\
\end{tabular}%
}
\end{table}

\subsection{Discussion}
\enlargethispage{1.5\baselineskip}

WandererPSO achieves the most consistent and scalable performance across all dimensionalities, whereas DrifterPSO exhibits the most polarized behavior and the poorest aggregate results. This contrast is particularly striking given the apparent similarity of their dynamics: both variants preserve the cognitive and social components of standard PSO. The crucial difference lies in how stochasticity is introduced. WandererPSO injects a scaled random vector into the velocity update, while DrifterPSO adds Gaussian noise directly to the position update. While this uncoupled positional disturbance allows particles to reliably ``teleport'' out of deceptive traps on highly rugged, multi-modal landscapes (e.g., Weierstrass), it completely destroys the swarm’s ability to conduct fine-grained exploitation on smoother topographies. When stochasticity is injected at the velocity level, exploratory movements remain regulated by inertia.

Among informed diversity mechanisms, negative learning strategies show differentiated effectiveness. DefeatistPSO achieves the highest number of best performances and strong aggregate results.
This approach guides a subswarm toward its personal worst positions, steering particles away from unproductive areas. Because it retains the global-best attractor, the algorithm maintains stable convergence.
Interestingly, applying this exact principle to the global worst position (ContrarianPSO) resulted in highly unreliable performance (18 worst-case outcomes). This performance gap suggests that
personal worst positions supply localized, trajectory-specific contextual information that may successfully bias the search toward unexplored, promising basins.

\textls[-1]{More broadly, the results indicate that preserving attraction toward the global best remains essential. Variants that retain the social component are the most successful algorithms in this study (WandererPSO, AmnesiacPSO, DefeatistPSO). Strategies that weaken, reverse, or ignore this component (such as RebelPSO or AnarchicPSO) generally fail to compete globally.
While the social term frequently triggers premature stagnation, eliminating it destabilizes the algorithm. It serves as a necessary anchor, enabling the swarm to collectively exploit a promising region once identified.
\enlargethispage{1.5\baselineskip}
}

\section{Conclusions}
\label{sec:conclusions}
\textls[-15]{The experimental results demonstrate that structured diversity enhancement through role-based subswarms significantly affects PSO performance in high-dimensional optimization.
Informed mechanisms are advantageous when they make effective use of locally meaningful failure memory. However, uninformed variants can perform just as well or better, especially when randomness is integrated into the velocity update while retaining both standard PSO components. This suggests that the method of injecting diversity is more important than whether it is problem-informed.
Hence, the effectiveness of diversity enhancement depends not only on the presence of stochasticity or counter-attraction mechanisms but rather on how they are integrated into swarm dynamics.
}

\begin{credits}
\subsubsection{\ackname} The computations were performed using resources provided by the Poznan Supercomputing and Networking Center (PCSS), Poland, within the HPC cluster Eagle under grant no. PL0590 (``Nature-inspired optimization'').

\smallskip
\noindent 
The research presented in this paper has been financially supported by: Polish National Science Center Grant no. 2019/35/O/ST6/00570 ``Socio-cognitive inspirations in classic metaheuristics.'';  Polish Ministry of Science and Higher Education funds assigned to AGH University of Science and Technology.
\end{credits}

\bibliographystyle{splncs04}
\bibliography{bibliography}

@book{bandura1986,
	author={Bandura, A.},
	title={Social learning theory},
	publisher={Prentice-Hall, Englewood Cliffs, N.J.},
	year={1971},
}

@article{kennedy1995particle,
  title={Particle swarm optimization},
  author={Kennedy, James and Eberhart, Russell},
  journal={Proceedings of ICNN'95-international conference on neural networks},
  volume={4},
  pages={1942--1948},
  year={1995},
  organization={IEEE}
}

@inproceedings{tbo,
author = {Urba\'{n}czyk, Aleksandra and Czech, Krzysztof and Urba\'{n}czyk, Piotr and Kisiel-Dorohinicki, Marek and Byrski, Aleksander},
title = {Socio-Cognitive Agent-Oriented Evolutionary Algorithm with Trust-Based Optimization},
year = {2025},
isbn = {978-3-031-97553-0},
publisher = {Springer-Verlag},
address = {Berlin, Heidelberg},
doi = {10.1007/978-3-031-97554-7\_18},
booktitle = {Computational Science – ICCS 2025 Workshops: 25th International Conference, Singapore, Singapore, July 7–9, 2025, Proceedings, Part I},
pages = {249–264},
numpages = {16},
location = {Singapore, Singapore}
}

@InProceedings{iccs2021,
author="Urba{\'{n}}czyk, Aleksandra
and Nowak, Bartosz
and Orzechowski, Patryk
and Moore, Jason H.
and Kisiel-Dorohinicki, Marek
and Byrski, Aleksander",
editor="Paszynski, Maciej
and Kranzlm{\"u}ller, Dieter
and Krzhizhanovskaya, Valeria V.
and Dongarra, Jack J.
and Sloot, Peter M. A.",
title="Socio-cognitive Evolution Strategies",
booktitle="Computational Science -- ICCS 2021",
year="2021",
publisher="Springer International Publishing",
address="Cham",
pages="329--342",
isbn="978-3-030-77964-1"
}

@article{urbanczykes2024,
title={($mu$ + $\lambda$) Evolution Strategy with Socio-cognitive Mutation}, volume={18},  
DOI={10.14313/JAMRIS/1-2024/1}, 
number={1}, 
journal={Journal of Automation, Mobile Robotics and Intelligent Systems}, 
author={Urbańczyk, Aleksandra and Kucaba, Krzysztof and Wojtulewicz, Mateusz and Kisiel-Dorohinicki, Marek and Rutkowski, Leszek and Duda, Piotr and Kacprzyk, Janusz and Yew Chong, Siang and Yao, Xin and Byrski, Aleksander}, 
year={2024}, 
pages={1–11} }

@article{eberhart1995new,
  title={A new optimizer using particle swarm theory},
  author={Eberhart, Russell C and Kennedy, James},
  journal={MHS'95. Proceedings of the Sixth International Symposium on Micro Machine and Human Science},
  pages={39--43},
  year={1995},
  organization={IEEE}
}

@article{shi1998modified,
  title={A modified particle swarm optimizer},
  author={Shi, Yuhui and Eberhart, Russell C},
  journal={1998 IEEE international conference on evolutionary computation proceedings. IEEE world congress on computational intelligence (Cat. No. 98TH8360)},
  pages={69--73},
  year={1998},
  organization={IEEE}
}

@inproceedings{kennedy1999small,
  title={Small worlds and mega-minds: Effects of neighborhood topology on particle swarm performance},
  author={Kennedy, James},
  booktitle={Proceedings of the 1999 congress on evolutionary computation-CEC99 (Cat. No. 99TH8406)},
  volume={3},
  pages={1931--1938},
  year={1999},
  organization={IEEE}
}

@Article{lopezibanez2016irace,
  author  = {Manuel López-Ibáñez and Jérémie Dubois-Lacoste and Leslie Pérez Cáceres and Mauro Birattari and Thomas Stützle},
  title   = {The irace Package: Iterated Racing for Automatic Algorithm Configuration},
  journal = {Operations Research Perspectives},
  volume  = {3},
  pages   = {43--58},
  year    = {2016},
  issn    = {2214-7160},
  doi     = {10.1016/j.orp.2016.09.002},
}

@Article{los2018misfit,
  author  = {Łoś, Marcin and Smołka, Maciej and Schaefer, Robert and Sawicki, Jakub},
  title   = {Misfit Landforms Imposed by Ill-Conditioned Inverse Parametric Problems\enlargethispage{2.5\baselineskip}},
  journal = {Computer Science},
  volume  = {19},
  number  = {2},
  pages   = {157},
  year    = {2018},
  doi     = {10.7494/csci.2018.19.2.2781},
}

@Article{benitezhidalgo2019jmetalpy,
  author  = {Antonio Benítez-Hidalgo and Antonio J. Nebro and José García-Nieto and Izaskun Oregi and Javier Del Ser},
  title   = {jMetalPy: A Python Framework for Multi-Objective Optimization with Metaheuristics},
  journal = {Swarm and Evolutionary Computation},
  volume  = {51},
  pages   = {100598},
  year    = {2019},
  issn    = {2210-6502},
  doi     = {10.1016/j.swevo.2019.100598},
}

@article{higashi2003gaussian,
  title={Gaussian distributed particle swarm optimization},
  author={Higashi, Norikazu and Iba, Hitoshi},
  journal={Proceedings of the 2003 IEEE Swarm Intelligence Symposium. SIS'03 (Cat. No. 03EX706)},
  pages={72--79},
  year={2003},
  organization={IEEE}
}

@article{esmin2013hybrid,
  title={A hybrid particle swarm optimization applied to loss power minimization},
  author={Esmin, A A A and Lambert-Torres, G and De Souza, A C Z},
  journal={IEEE transactions on power systems},
  volume={20},
  number={2},
  pages={859--866},
  year={2005},
  publisher={IEEE}
}

@article{secrest2003gaussian,
  title={Gaussian particle swarm optimization},
  author={Secrest, Bruce R and Lamont, Gary B},
  journal={Proceedings of the 2003 ACM symposium on Applied computing},
  pages={1213--1217},
  year={2003}
}

@article{zhang2014adaptive,
  title={Adaptive Gaussian disturbance based bare-bones particle swarm optimization},
  author={Zhang, Yin and Li, Shangce and Wang, Ling and Wang, Jun},
  journal={Soft Computing},
  volume={18},
  number={12},
  pages={2407--2420},
  year={2014},
  publisher={Springer}
}

@article{stacey2003mutation,
  title={Mutation in particle swarm optimization},
  author={Stacey, A and Jancic, M and Grundy, I},
  journal={2003 IEEE Swarm Intelligence Symposium. SIS'03 (Cat. No. 03EX706)},
  pages={128--133},
  year={2003},
  organization={IEEE}
}

@article{wang2007cauchy,
  title={A new particle swarm optimization with Cauchy mutation},
  author={Wang, Ling and Li, Shangce},
  journal={2007 IEEE Swarm Intelligence Symposium},
  pages={145--149},
  year={2007},
  organization={IEEE}
}

@article{krohling2009barebones,
  title={Bare bones particle swarm optimization with Gaussian or Cauchy jumps},
  author={Krohling, Renato A and Mendel, Jerry M},
  journal={2009 IEEE Congress on Evolutionary Computation},
  pages={245--252},
  year={2009},
  organization={IEEE}
}

@article{hakli2014levy,
  title={A novel particle swarm optimization algorithm with L{\'e}vy flight},
  author={Hakli, H and U{\u{g}}uz, H},
  journal={Applied Soft Computing},
  volume={23},
  pages={333--345},
  year={2014},
  publisher={Elsevier}
}

@article{yong2015adaptive,
  title={An adaptive particle swarm optimization with chaotic mapping and perturbation},
  author={Yong, Wang and Jian-Zhong, Zhou and Hui, Li and Hui-Juan, Lu},
  journal={Applied Mathematics and Computation},
  volume={268},
  pages={1083--1097},
  year={2015},
  publisher={Elsevier}
}

@article{chen2021adaptive,
  title={An adaptive Gaussian mutation particle swarm optimization algorithm},
  author={Chen, Kai and Zhou, Fuhai and Wang, Ling and Liu, Feng},
  journal={Applied Mathematical Modelling},
  volume={90},
  pages={743--761},
  year={2021},
  publisher={Elsevier}
}

@article{xu2025hybrid,
  title={A hybrid differential evolution and particle swarm optimization algorithm with adaptive strategy selection},
  author={Xu, Jin and Liu, Jing and Li, Yun},
  journal={Soft Computing},
  volume={29},
  number={1},
  pages={1--20},
  year={2025},
  publisher={Springer}
}

@article{zhang2015comprehensive,
  title={A comprehensive survey on particle swarm optimization algorithm and its applications},
  author={Zhang, Y and Wang, S and Ji, G},
  journal={Mathematical Problems in Engineering},
  volume={2015},
  year={2015},
  publisher={Hindawi}
}

@article{liang2013cec2013,
  title={Problem definitions and evaluation criteria for the CEC 2013 special session on real-parameter optimization\enlargethispage{2.5\baselineskip}},
  author={Liang, JJ and Qu, BY and Suganthan, PN},
  journal={Technical Report 201212, Computational Intelligence Laboratory, Zhengzhou University, Zhengzhou, China and Nanyang Technological University, Singapore},
  volume={635},
  year={2013}
}

@article{wu2017cec2017,
  title={Problem definitions and evaluation criteria for the CEC 2017 special session and competition on single objective real-parameter numerical optimization},
  author={Wu, G and Mallipeddi, R and Suganthan, PN},
  journal={Technical Report, Nanyang Technological University, Singapore},
  year={2016}
}

@article{jamil2013survey,
  title={A literature survey of benchmark functions for global optimization problems},
  author={Jamil, Momin and Yang, Xin-She},
  journal={International Journal of Mathematical Modelling and Numerical Optimisation},
  volume={4},
  number={2},
  pages={150--194},
  year={2013},
  publisher={Inderscience Publishers}
}

@article{piotrowski2023benchmark,
  title={Benchmark functions for the CEC 2022 competition on single objective bound constrained numerical optimization},
  author={Piotrowski, Adam P and Nobile, Mario S and Suganthan, Ponnuthurai N and B{\l}achowicz, Tomasz and Cazzaniga, Paolo and D\k{a}browski, Marcin and Lang, Kevin and Lango, Mateusz and Mykhailov, Vitalii and Piotrowska, Anna and others},
  journal={IEEE Transactions on Evolutionary Computation},
  year={2023},
  publisher={IEEE}
}

@inproceedings{urbanczyk2025sequential,
  author    = {Piotr Urbańczyk and Aleksandra Urbańczyk and Magdalena Król and Leszek Rutkowski and Marek Kisiel-Dorohinicki},
  title     = {Sequential, parallel and consecutive hybrid evolutionary-swarm optimization metaheuristics},
  booktitle = {International Conference on Computational Science (ICCS)},
  year      = {2025},
  publisher = {Springer},
  series    = {Lecture Notes in Computer Science},
  doi       = {10.1007/978-3-031-97554-7_15},
}

@article{abualigah2025particle,
  author    = {Laith Abualigah},
  title     = {Particle Swarm Optimization: Advances, Applications, and Experimental Insights\enlargethispage{2.5\baselineskip}},
  journal   = {Computers, Materials \& Continua},
  year      = {2025},
  volume    = {82},
  number    = {2},
  pages     = {1539--1592},
  doi       = {10.32604/cmc.2025.060765},
  publisher = {Tech Science Press}
}

@article{thangaraj2011particle,
  author = {Radha Thangaraj and Millie Pant and Ajith Abraham and Pascal Bouvry},
  title = {Particle swarm optimization: Hybridization perspectives and experimental illustrations},
  journal = {Applied Mathematics and Computation},
  volume = {217},
  number = {12},
  pages = {5208--5226},
  year = {2011},
  issn = {0096-3003},
  doi = {10.1016/j.amc.2010.12.053}
}

@article{sengupta2019particle,
  author = {Saptarshi Sengupta and Sanchita Basak and Richard Alan Peters},
  title = {Particle Swarm Optimization: A Survey of Historical and Recent Developments with Hybridization Perspectives},
  journal = {Machine Learning and Knowledge Extraction},
  volume = {1},
  number = {1},
  pages = {157--191},
  year = {2019},
  issn = {2504-4990},
  doi = {10.3390/make1010010}
}

@article{jain2022Overview,
  author = {Meetu Jain and Vibha Saihjpal and Narinder Singh and Satya Bir Singh},
  title = {An Overview of Variants and Advancements of PSO Algorithm},
  journal = {Applied Sciences},
  volume = {12},
  number = {17},
  pages = {8392},
  year = {2022},
  issn = {2076-3417},
  doi = {10.3390/app12178392}
}

@book{minku2023introduction,
  title     = {Introduction to Computational Intelligence},
  editor    = {Minku, Leandro L. and Cabral, George and Martins, Marcella and Wagner, Markus},
  publisher = {IEEE Computational Intelligence Society},
  year      = {2023},
  doi       = {10.5281/zenodo.7537827},
  note      = {An IEEE Computational Intelligence Society Open Book}
}

@misc{chauhan2025learning,
  author       = {Chauhan, Dikshit and Shivani and Suganthan, P. N.},
  title        = {Learning Strategies in Particle Swarm Optimizer: A Critical Review and Performance Analysis},
  year         = {2025},
  url          = {https://arxiv.org/abs/2504.11812}
}

@Misc{riget2002arpso,
  author    = {Riget, Jacques and Vesterstrøm, Jakob S.},
  title     = {A Diversity-Guided Particle Swarm Optimizer --- the ARPSO},
  howpublished = {EVALife Technical Report no. 2002-02, Department of Computer Science, Aarhus University},
  address   = {Aarhus, Denmark},
  year      = {2002},
  pages     = {1--13}
}

@Article{alkazemi2000multiphase,
  author  = {Al-kazemi, Buthainah Sabeeh No'man and Mohan, Chilukuri K.},
  title   = {Multi-phase Discrete Particle Swarm Optimization},
  journal = {Electrical Engineering and Computer Science - All Scholarship},
  volume  = {54},
  year    = {2000},
  publisher = {Syracuse University},
  url     = {https://surface.syr.edu/eecs/54}
}

@InProceedings{alkazemi2002multiphase,
  author    = {Al-kazemi, B. and Mohan, C.K.},
  title     = {Multi-phase generalization of the particle swarm optimization algorithm},
  booktitle = {Proceedings of the 2002 Congress on Evolutionary Computation. CEC'02 (Cat. No.02TH8600)},
  year      = {2002},
  volume    = {1},
  pages     = {489--494},
  doi       = {10.1109/CEC.2002.1006283}
}

@InProceedings{chen2005twosubswarms,
  author    = {Chen, Guochu and Yu, Jinshou},
  editor    = {Wang, Lipo and Chen, Ke and Ong, Yew Soon},
  title     = {Two Sub-swarms Particle Swarm Optimization Algorithm},
  booktitle = {Advances in Natural Computation},
  year      = {2005},
  publisher = {Springer Berlin Heidelberg},
  address   = {Berlin, Heidelberg},
  pages     = {515--524},
  isbn      = {978-3-540-31863-7},
  doi       = {10.1007/11539902_63}
}

@Article{xia2014reverselearning,
  author  = {Xuewen Xia and Jingnan Liu and Yuanxiang Li},
  title   = {Particle Swarm Optimization Algorithm with Reverse-Learning and Local-Learning Behavior},
  journal = {Journal of Software},
  volume  = {9},
  number  = {2},
  pages   = {350--357},
  year    = {2014},
  doi     = {10.4304/jsw.9.2.350-257}
}

@InProceedings{dong2018reverselearning,
  author    = {Dong, Hongbin and Zhang, Hua and Han, Shuang and Li, Xiaohui and Wang, Xiaowei},
  title     = {Reverse-Learning Particle Swarm Optimization Algorithm Based on Niching Technology},
  booktitle = {2018 IEEE/ACIS 17th International Conference on Computer and Information Science (ICIS)},
  year      = {2018},
  pages     = {405--410},
  doi       = {10.1109/ICIS.2018.8466391}
}

@Article{comak2016generalized,
  author  = {Çomak, Emre},
  title   = {A Generalized Particle Swarm Optimization Using Reverse Direction Information},
  journal = {Turkish Journal of Electrical Engineering and Computer Sciences},
  volume  = {24},
  number  = {2},
  pages   = {Article 22},
  year    = {2016},
  doi     = {10.3906/elk-1306-39},
}

@Article{mason2016avoidance,
  author  = {Mason, Karl and Howley, Enda},
  title   = {Exploring Avoidance Strategies \& Neighbourhood Topologies in Particle Swarm Optimisation},
  journal = {International Journal of Swarm Intelligence},
  volume  = {2},
  year    = {2016},
  doi     = {10.1504/IJSI.2016.10002177}
}

@Article{kumar2023parallel,
  author  = {Lalit Kumar and Manish Pandey and Mitul Kumar Ahirwal},
  title   = {Parallel Global Best-Worst Particle Swarm Optimization Algorithm for Solving Optimization Problems},
  journal = {Applied Soft Computing},
  volume  = {142},
  pages   = {110329},
  year    = {2023},
  issn    = {1568-4946},
  doi     = {10.1016/j.asoc.2023.110329},
}

\end{document}